\pdfoutput=1

\documentclass[11pt]{article}

\usepackage{acl}

\usepackage{times}
\usepackage{amsthm}
\usepackage{latexsym}
\usepackage{graphicx}
\usepackage{color}
\usepackage{caption}
\usepackage{subcaption}
\usepackage{bbm}
\usepackage{url}
\usepackage{amssymb}
\usepackage{amsmath}
\usepackage{multirow}
\usepackage{booktabs}
\usepackage{algorithm}
\usepackage{algpseudocode}
\usepackage{wrapfig}

\newcommand{\algo}{\textsc{DAdEE}}

\title{\algo{}: Unsupervised Domain Adaptation in Early Exit PLMs}

\author{Divya Jyoti Bajpai and Manjesh Kumar Hanawal\\
   Department of IEOR, IIT Bombay \\
  \texttt{\{divyajyoti.bajpai, mhanawal\}@iitb.ac.in}}

\begin{document}
\maketitle

\begin{abstract}

Pre-trained Language Models (PLMs) exhibit good accuracy and generalization ability across various tasks using self-supervision, but their large size results in high inference latency. Early Exit (EE) strategies handle the issue by allowing the samples to exit from classifiers attached to the intermediary layers, but they do not generalize well, as exit classifiers can be sensitive to domain changes. To address this, we propose Unsupervised Domain Adaptation in EE framework (\algo{}) that employs multi-level adaptation using \textit{knowledge distillation}. \algo{} utilizes GAN-based adversarial adaptation at each layer to achieve domain-invariant representations, reducing the domain gap between the source and target domain across all layers. The attached exits not only speed up inference but also enhance domain adaptation by reducing catastrophic forgetting and mode collapse, making it more suitable for real-world scenarios. Experiments on tasks such as sentiment analysis, entailment classification, and natural language inference demonstrate that \algo{} consistently outperforms not only early exit methods but also various domain adaptation methods under domain shift scenarios. The anonymized source code is available at \url{https://github.com/Div290/DAdEE}.

\end{abstract}


\section{Introduction}
Pre-trained Language Models (PLMs) such as BERT \cite{devlin2018bert}, GPT \cite{radford2019language}, XLNet \cite{yang2019xlnet}, and RoBERTa \cite{liu2019roberta} have experienced substantial growth in size to attain state-of-the-art performances across a wide range of natural language processing tasks. Despite their remarkable efficacy, PLMs suffer from inference latencies, limiting their utility in industrial applications requiring faster inference. Prior research \cite{zhou2020bert, zhu2021leebert} has also highlighted overthinking issues in PLMs. More specifically, shallow representations in the initial layers may suffice for correct inference of `easy' samples, while final layer representations may overfit or be influenced by irrelevant features, resulting in poor generalization and computational inefficiency.

 To circumvent this, several methods such as pruning \cite{michel2019sixteen}, quantization \cite{kim2021bert} and knowledge distillation \cite{jiao2019tinybert} have been proposed. Among them, Early Exit methods \cite{zhou2020bert, zhu2021leebert} have gained significant attention, where inference can be made at classifiers attached at intermediary layer based on the hardness of the input samples. They perform adaptive inference strategies to address both efficiency and overthinking concerns.
 
Nevertheless, all these methods do not generalize well to new domains. Also, as EE introduces more parameters in the exit classifers, it requires high-quality labeled training data to learn the exit weights. The cost of creating labeled training data is often prohibitively expensive. The question then arises: How can we adapt early exit PLMs to diverse domains in an unsupervised setup?



\noindent

{ \it Domain adaptation} is a vital technique in machine learning to ensure models perform well on data from a target domain, even if trained on a different source domain. Unsupervised domain adaptation methods primarily focus on aligning source and target data in a shared feature space. This alignment is achieved by optimizing the model's representation to minimize domain discrepancies, such as maximum mean discrepancy \cite{tzeng2014deep} or correlation instances \cite{sun2016deep}. Previous approaches, like adversarial training methods \cite{ajakan2014domain, tzeng2017adversarial, ryu2022knowledge}, have utilized adversarial objectives to bridge domain gaps by producing domain-invariant features at the final layer. However, since the Early Exit PLMs (EEPLMs)  have added exits, domain adaptation methods cannot be directly used as they only focus on adapting to the final layer performance. This necessitates domain adaptation at every layer such that domain invariant features are not only available at the final layer but across all the layers of PLM.

To tackle this problem, we present a novel strategy that integrates adversarial domain adaptation techniques in early exits that adapt EEPLMs to diverse domains. Our approach, named Unsupervised \underline{D}omain \underline{Ad}aptation in \underline{EE} PLMs (\algo), emphasizes achieving domain invariant representations across all the layers such that exit classifiers trained on a source domain can be directly utilized for the target domain. This allows for the exit classifiers trained on the source domain to be directly utilized in the target domain without requiring labels.

Our method not only speeds up inference but also allows for fast and robust bridging of domain gaps compared to traditional methods that rely solely on final layer representations in adversarial setups, which is insufficient for such large and complex models. During adversarial training, the exits aid the adaptive process by mitigating the risk of catastrophic forgetting and mode collapse using knowledge distillation between source and target representations across all the layers. 


By leveraging multi-level domain feature adaptation, \algo{} enhances overall effectiveness across real-world applications, addressing the challenges of inference speed and domain adaptation in PLMs. Also, in scenarios where the size of the source dataset is limited, early exit models can adapt more readily as these models do not depend on the prediction of just the final layer but multiple layers, improving generalization.
Our method achieves improved performance metrics on sentiment analysis, entailment classification, and natural language inference (NLI) tasks.

\begin{figure*}
    \centering
    \includegraphics[scale = 0.55]{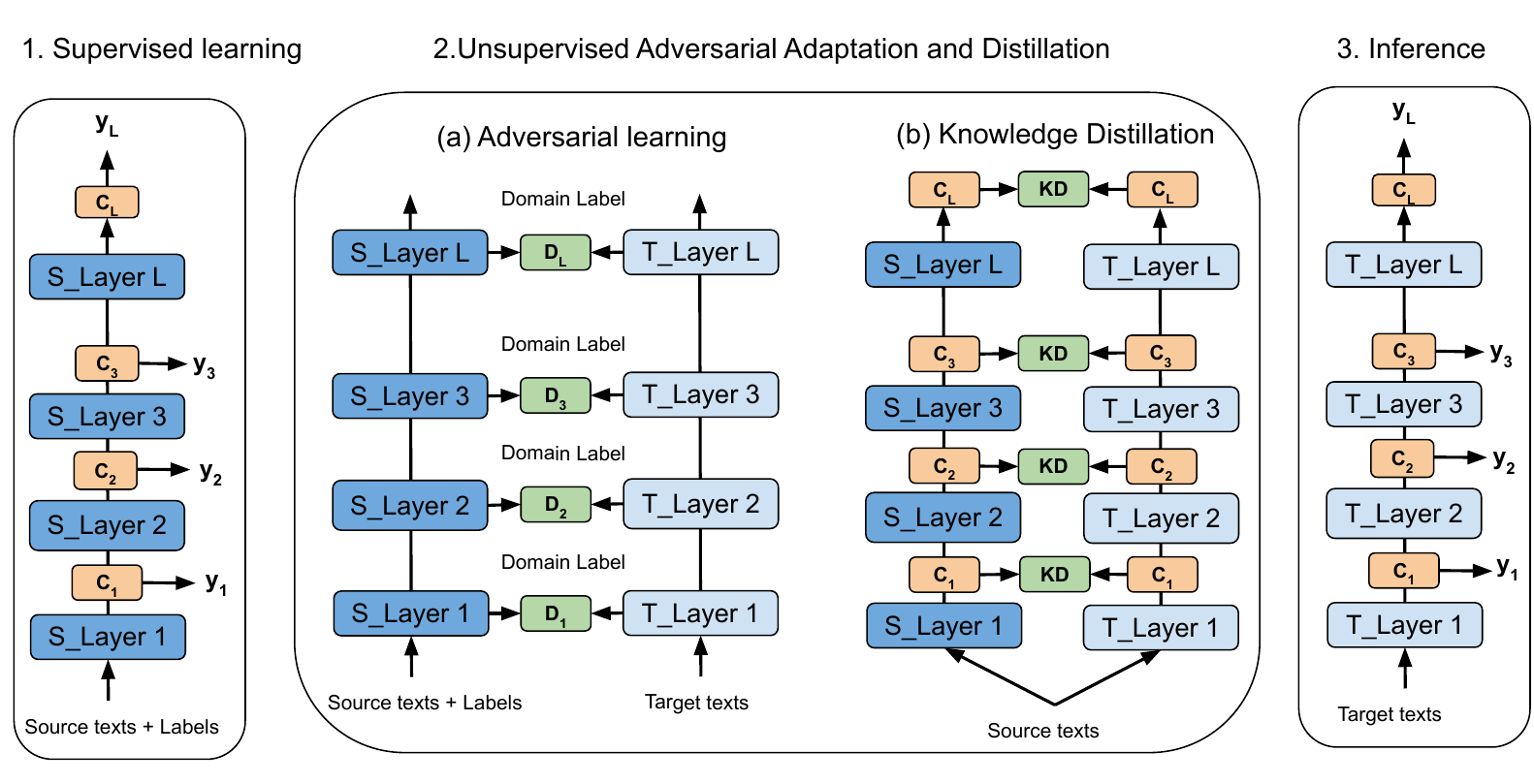}
    \caption{
This figure outlines our model's workflow with three main steps: 1) Supervised learning with attached exits during backbone training. 2) Adversarial adaptation and distillation, where a) a discriminator predicts domain labels at each layer, and b) the target backbone is trained for domain-invariant features. 3) The target backbone produces domain-invariant features across all layers, ready for target domain inference.}
    \label{fig:main}
\end{figure*}

Our main contributions are as follows:
\begin{itemize}
\item We propose \algo{} for unsupervised domain adaptation to bridge the gaps between the source and target domains in EEPLMs.

\item \algo{} uses a GAN-based multi-level adaptation to bridge the domain gaps, i.e., we perform layer-by-layer adaptation. 

\item We utilize EE strategies not only for faster inference but also for the adaptation process. Our method gets the best of both early exit and domain adaptation methods to simultaneously increase both performance and speed.

\item Through extensive experiments on sentiment analysis, entailment classification and natural language inference (NLI) tasks, we show that \algo{} achieved an average improvement of $2.9\%$ in accuracy and $1.61\times $ average inference speed up as compared to previous vanilla PLM inference. 
\end{itemize}

\section{Related works}
In this section, we discuss studies relevant to our work in domain adaptation and early exiting.

\noindent
\textbf{Domain Adaptation:} The aim is to learn domain-invariant representations for labeled source and unlabeled target domains. Key methods include Deep Domain Classification \cite{tzeng2014deep}, which minimizes maximum mean discrepancy with classification loss, and Deep Adaptation Network \cite{long2015learning}, employing multiple kernels across layers. DCA \cite{sun2016deep} reduces the disparity in second-order statistics. Adversarial methods like DANN \cite{ajakan2014domain} use a domain classifier with a gradient reversal layer for domain confusion. Similarly Domain Separation Networks (DSN) \cite{bousmalis2016domain} have a notion of private subspaces for every domain and separate the information for each domain.

Generative approaches, such as CyCADA \cite{hoffman2018cycada}, enforce cycle and semantic consistency. DAAT \cite{du2020adversarial} enhances domain awareness through post-training, while ADDA \cite{tzeng2017adversarial} introduces GAN-based loss, further improved by AAD \cite{ryu2022knowledge} with knowledge distillation. Pivot-based methods \cite{blitzer2007biographies, yu2016learning, ziser2018pivot, peng2018cross, zhang2019interactive} induce shared low-dimensional features based on pivot co-occurrence. Multi-level domain adaptation methods \cite{malik2023udapter} utilize all layers to bridge domain gaps.

\noindent
\textbf{Early Exits:} are input adaptive inference methods. For image classification, BranchyNet \cite{teerapittayanon2016branchynet} uses classification entropy for early inference, while MSDNet \cite{huang2017multi} selects thresholds based on confidence distribution. DeeCAP \cite{fei2022deecap} and MuE \cite{tang2023you} extend it to image captioning.

Early exiting has also been applied to PLMs for various NLP tasks \cite{bapna2020controlling, elbayad19arxiv, liu2021elasticbert,xin2021berxit, zhou2020bert, he2021magic, banino2021pondernet, ji2023early}. DeeBERT \cite{xin2020deebert} and ElasticBERT \cite{liu2021elasticbert} propose different fine-tuning strategies. PABEE \cite{zhou2020bert} bases early exit decisions on prediction consistency. DDA \cite{li2021dynamic} uses exits for dynamic domain adaptation. It considers a feature classifier to bridge the domain gap which is not sufficient when there is a larger domain gap. CeeBERT \cite{bajpai2024ceebert} adapts thresholds for target domains using multi-armed bandits.

The key differences in our work are: 1) We are the first to perform domain adaptation in the early exit PLMs.
2) We utilize a GAN-based framework for adversarial training across all exits for multi-level domain adaptation. 3) Our exits employ knowledge distillation between the source and target domains to mitigate catastrophic forgetting, thus outperforming both the domain adaptation and early exit methods.


\section{Methodology}
In this section, we detail our method. We start with a PLM such as BERT/RoBERTa with $L$ layers. We attach exit classifiers to each layer and train them using the source dataset. We then perform multi-level adversarial training {on target data}.
\subsection{Training the source backbone}
For any source sample $(x_s, y_s)$, 
the loss at exit classifier $i$ is computed as:
\begin{equation}
    \mathcal{L}_i(\theta) = \mathcal{L}_{CE}(f_i(x_s, \theta), y_s),
\end{equation}
where $f_i(x_s, \theta)$ is the output of the classifier attached at the $i$th layer, $\theta$ is the set of learnable parameters of the source backbone, and $\mathcal{L}_{CE}$ is the cross-entropy loss. We learn the parameters for all the exit-classifiers simultaneously following the approach of Shallow-Deep \cite{kaya2019shallow}, with the loss function defined as $\mathcal{L} = \frac{\sum_{i = 1}^{L} i\cdot\mathcal{L}i}{\sum_{i = 1}^{L} i}$, over all the $L$ layers. This weighted average considers the relative inference cost of each internal classifier. This is shown as the step Supervised learning in Fig~\ref{fig:main}. After this training, the weights of the source encoder and exit classifiers at each layer are frozen. We note that as a part of the training, a function $C_i$ that maps the source encoder's output to class probabilities at $i$th layer is also learned and frozen.  


\subsection{Adversarial adaptation on target}
We initialize the target encoder weights with what is learned for the source encoder. We assume that the label space of the target dataset is the same as the source label space $\mathcal{C}$. We denote outputs of the $i$th layer of the source and target encoder as $E_i^s(x)$ and $E_i^t(x)$, respectively. Let $D_i$ denote a discriminator function that maps the output of $i$th layer of either source encoder or target encoder to domain probabilities, i.e., target or source domain.

We train the target encoder and the discriminator alternately as the ADDA framework \cite{tzeng2017adversarial}. This can be formulated as an unconstrained optimization problem and is represented as step 2(a): Adversarial Framework in Fig.~ \ref{fig:main}. Let $D_s$ and $D_t$ denote the target distributions. For $x_s\sim D_s$ and $x_t\sim D_t$, loss for $i$th discriminator can be formulated as
\vspace{-.2cm}
\begin{multline}
    \mathcal{L}_i^{dis}(x_s, x_t) = -\text{log}D_{i}(E_{i}^{s}(x_s))
    \\- \text{log}(1-D_i(E_i^{t}(x_t))),
\end{multline}
and the overall discriminator loss across all the layers can be as $\mathcal{L}^{dis} = \frac{\sum_{i = 1}^{L} i\cdot\mathcal{L}_i^{dis}}{\sum_{i = 1}^{L} i}$. The generator loss for $i$th layer is:
\begin{equation}
    \mathcal{L}_{i}^{gen}(x_t) = -\text{log}D_i(E_i^{t}(x_t))
\end{equation}
Given that the weights of the target encoder are untied from those of the source encoder, the target encoder has more flexibility in learning the specific domain features. However, this formulation is prone to catastrophic forgetting \cite{ryu2022knowledge}, leading to erratic classification performance, as it lacks access to class labels and can diverge from the original task.

To address this challenge, we employ knowledge distillation. This involves introducing distillation loss alongside generator loss after each layer to mitigate the risk of catastrophic forgetting and mode collapse across all layers. We ensure robustness throughout the model by leveraging classifiers (exits) appended after each layer to distil the knowledge between source and target. The formulation for knowledge distillation loss is expressed as follows:
\begin{equation}
    \mathcal{L}_i^{KD} = KL(C_i(E_i^s(x_s)), C_i(E_i^t(x_s)))
\end{equation}
where 
$KL$ is the KL-divergence loss. The total loss of generator of the $i$th classifier then becomes $\mathcal{L}_i = \mathcal{L}_i^{gen}+\mathcal{L}_i^{KD}$, and the overall generator loss is taken as $\mathcal{L} = \frac{\sum_{i = 1}^{L} i\cdot\mathcal{L}_i}{\sum_{i = 1}^{L} i}$. We provide lower weights to the discriminator and generator at initial layers since these layers are responsible for learning the general features which should not be changed much while deeper domain-specific rich representations lie in deeper layers justifying higher weights to deeper layers.

After this step, the target backbone is ready for inference as discussed next.

\subsection{Inference on target dataset}
For any $x_{t}\sim \mathcal{D}_t$, let $p_i^{t}(c)$ denote the probability assigned at layer $i$ that a sample belongs to class $c\in \mathcal{C}$. 
 Let $S_i:= \max_{c\in \mathcal{C}}p_i^{t}(c)$. It denotes the confidence in prediction at the $i$ layer. 
The decision to exit early is made based on this confidence score exceeding a fixed threshold $\alpha$, i.e., a sample exit from layer $i$ if $S_i\geq \alpha$, else it is processed to the next layer. When the sample exits at layer $i$, it is assigned a label $\arg\max_{c \in \mathcal{C}} p^t_i (c)$.
If the sample's confidence is below $\alpha$ for all the intermediary layers, the sample is inferred at the final layer.

We set the value of $\alpha$ using the validation split of the source dataset. We use the same threshold for the target dataset as after adversarial training as all the layers provide domain invariant feature representation resulting in a similar accuracy-efficiency trade-off in the target domain as learned on the source domain.
\subsection{Analysis}
This section provides a theoretical justification of \algo{}. Initially, we delve into the existing theoretical framework, subsequently elucidating the advantageous aspects of early exits and adversarial training from a theoretical standpoint. We follow the method pioneered in \cite{ben2010theory} to upper bound expected error of a hypothesis on the target domain. 

Let $h_s^{*}$ and $h_t^{*}$ denote the hypotheses that assign ground-truth labels for the source and target domain, respectively. 
We define the disagreement function  for any hypothesis $h_1$ and $h_2$ as :
\begin{equation}
\epsilon(h_1, h_2) = \mathbb{E}[|h_1(x)-h_2(x)|].
\end{equation} 
For any hypothesis $h$, define $\epsilon_s(h) = \epsilon_s(h, h_t^{*})$ and $\epsilon_t(h) = \epsilon_t(h, h_s^{*})$ as the expected error on the source and target domain respectively. 
The error $\epsilon_t(h)$ of a hypothesis $h$ on the target domain can be bounded using three terms: (a) expected error of $h$ on the source domain, $\epsilon_s(h)$; (b) $\mathcal{H}\Delta\mathcal{H}$-distance $d_{\mathcal{H}\Delta\mathcal{H}}(\mathcal{D}_s, \mathcal{D}_t)$ measuring domain shift as the discrepancy between the disagreement of the two hypotheses $h, h^{'}\in \mathcal{H}$ which is defined as 
\begin{equation}\label{eq: distance}
d_{\mathcal{H}\Delta\mathcal{H}}(\mathcal{D}_s, \mathcal{D}_t) = 2\sup_{h, h^{'}\in \mathcal{H}}|\epsilon_s(h,h^{'})-\epsilon_t(h, h^{'})|
\end{equation}
and (c) the error $\lambda$ of the ideal joint hypothesis $h^{*}$ on both source and target domains. The upper bound is:
\begin{equation}
    \epsilon_t(h)\leq \epsilon_{s}(h)+\frac{1}{2}d_{\mathcal{H}\Delta\mathcal{H}}+\lambda
\end{equation}
Usually, $\lambda$ is considered negligible and discarded. Therefore, we can focus on making the first and second terms small to keep the target error small. 

For the first term, the error rate of the source domain is minimized by training it on the labeled training data. For the second term, it is required that the PLM generate similar features for the source as well as the target dataset. By imposing multi-level adaptation, we reduce the value of the second term in the upper bound. In our method, all the exits simultaneously bridge the domain gap across all the layers instead of just the final one. This helps in layer-by-layer adaptation to the target domain, producing similar feature representations for the target and source datasets across all the layers instead of just the final layer. This helps significantly reduce the second term. We demonstrate it in figure \ref{fig:A-distance} by calculating the $\mathcal{A}$-distance. 
Note that if knowledge distillation is not applied at every layer, it could increase the value of the first term because of mode collapse or catastrophic forgetting. This aspect was not considered in previous methods like \cite{ryu2022knowledge}, where knowledge distillation is applied only at the last layer. 
\section{Experiments}
In this section, we elaborate on all the experimental details of our work.

\subsection{Datasets}
We conduct experiments on well-established benchmark datasets sourced from Amazon reviews \cite{blitzer2007biographies}. These datasets cover reviews from four distinct domains: Books (B), DVDs (D), Electronics (E), and Kitchen appliances (K). Additionally, we utilize the Airline review dataset (A) \cite{nguyen2015airline} and the IMDB dataset (I) \cite{maas2011learning}. In total, our study encompasses 30 domain adaptation tasks for sentiment analysis. For NLI and entailment classification, we use the datasets available in GLUE \cite{wang2019glue} and ELUE \cite{liu2021elasticbert} tasks. 

For each domain, we use the train split of the source (with labels) and the target domain (without labels) to train and adapt the backbone. The validation split of the source dataset is used for development and then finally the model is tested on the target test set.


\subsection{Implementation details}
\textbf{Training:} Initially, we train the backbone on the source dataset. We add a linear classifier layer after each intermediate layer of the pre-trained BERT/RoBERTa model, running the model for three epochs. The training uses a batch size of 16 and a learning rate of 1e-5 with the ADAM optimizer \cite{kingma2014adam}. We apply early stopping and select the best-performing model on the development set. Subsequently, we freeze the source encoder and initialize the target encoder with the source encoder's weights. The discriminator is an MLP with two hidden layers (hidden size 3072) and LeakyReLU activation. The domain adaptation step runs for five epochs with the same hyperparameters. The experiments are conducted on a single NVIDIA RTX 2070 GPU with an average runtime of less than 10 minutes.

\subsection{Inference} 
During inference, we use a batch size of 1 and $\alpha$ is chosen as the best-performing threshold on the source dataset's validation split based on accuracy. The search space for $\alpha$ is $S_{\alpha} = \{0.8, 0.85, 0.9, 0.95, 1.0\}$. We apply the same threshold as learned on the source dataset since all layers produce domain-invariant features, allowing the threshold to work effectively for the target domain as well.

\textbf{Speedup metric:} To maintain consistency with previous methods, we use the speedup ratio as the metric to assess our model which could be written as: 
$\frac{\sum_{i = 1}^L L\times n_i}{\sum_{i = 1}^L i\times n_i}$
where $n_i$ are the number of samples exiting from the $i$th layer and $L$ represents the number of layers. This metric could be interpreted as the increase in speed of the model as compared to the naive BERT/RoBERTa models.

\subsection{Baselines}
We compare our method against state-of-the-art early exiting and domain adaptation techniques, categorizing the baselines into four groups:

\textbf{1) Backbone:} We use vanilla BERT/RoBERTa as the primary baseline, in this we fine-tune the model on source domain data and infer on complete target data without adaptation.

\textbf{2) Domain Adaptation:} This category includes methods focused solely on domain adaptation.
DANN \cite{ajakan2014domain} employs adversarial training on deep neural networks, adapting the representation encoded in a 5000-dimensional feature vector of frequent unigrams and bigrams.
IATN \cite{zhang2019interactive} features two attention networks: one identifies common features between domains through domain classification, and the other extracts information from aspects using these common features.
DAAT \cite{du2020adversarial} initializes with BERT’s pre-trained weights and adapts it using novel self-supervised pre-training tasks followed by adversarial training. This method is specific to BERT and not easily extendable to other PLMs.
AAD \cite{ryu2022knowledge} uses adversarial training with distillation from the source to reduce overfitting.
UDApter \cite{malik2023udapter} performs unsupervised multi-level domain adaptation.

\textbf{3) Early Exit Methods:}
PABEE \cite{zhou2020bert} is a state-of-the-art early exit method that exits based on prediction consistency. This is chosen as a baseline to give an interpretation of how vanilla early exiting performs without adaptation.

\textbf{4) Domain Adaptation with Early Exit:}
DDA \cite{li2021dynamic} performs multi-level adaptation for image datasets to enable faster inference, though it can suffer from catastrophic forgetting.
CeeBERT \cite{bajpai2024ceebert} uses multi-armed bandits to adapt thresholds, focusing on inference efficiency without directly addressing the domain gap.
In the result tables, `Final' represents the performance of our method when inference is conducted solely at the final layer after adaptation, without utilizing early exits.
\begin{table*}[ht]
\centering
\small
\begin{tabular}{c|c|ccccc|ccc|cc}
\hline
& & \multicolumn{5}{c|}{\textit{Domain Adaptation methods}}& \multicolumn{3}{c|}{\it Early Exit models} &\\
\textbf{S $\rightarrow$ T}  & \textbf{BERT} & \textbf{DANN} & \textbf{IATN} & \textbf{DAAT} & \textbf{AAD} & \textbf{UDA} & \textbf{PABEE} & \textbf{CBRT} &\textbf{DDA} &\textbf{Final} & \textbf{Our}  \\

\hline
B $\rightarrow$ D           & 85.9          & 85.5          & 86.3          & 87.2         & 86.6      &  87.5  & 86.0      &  86.5    & 86.9 & 88.5           & \textbf{88.7} \\
B $\rightarrow$ E           & 85.7          & 84.3          & 86.5          & 87.5         & 86.7    &    87.3  & 85.3       & 86.3   & 87.1 & 87.4           & \textbf{87.8} \\
B $\rightarrow$ K           & 87.4          & 86.7          & 88.1          & 89.2         & 88.3    &  88.9  & 87.1     &   89.4   & 89.6 & 89.8           & \textbf{90.6} \\
B $\rightarrow$ A           & 84.5          & 83.5          & 85.3          & 86.5         & 85.9    &   86.8   & 84.9     &   86.0   & 86.3 & 86.7           & \textbf{87.1} \\
B $\rightarrow$ I           & 83.1          & 82.9          & 83.6          & 83.9         & 82.7    &   84.1   & 83.3      &  83.5   & 83.8 & 83.6           & \textbf{84.0} \\
D $\rightarrow$ B           & 85.0          & 83.7          & 86.6          & 87.4         & 87.0    &   87.2   & 84.5     &   86.9   & 88.1 & 87.9           & \textbf{88.4} \\
D $\rightarrow$ E           & 83.9          & 81.9          & 84.2          & 86.0         & 85.6    &   85.9   & 84.2    &    85.8   & 86.4 & 86.3           & \textbf{86.5} \\
D $\rightarrow$ K           & 85.3          & 85.5          & 85.8          & 87.8         & 87.4     &   88.0  & 84.7  &    86.1     & 87.8 & 88.1           & \textbf{88.5} \\
D $\rightarrow$ A           & 80.7          & 81.9          & 82.6          & \textbf{86.7}         & 84.1   &     86.4   & 80.3      &  85.2   & 86.3 & 86.5           & 86.6 \\
D $\rightarrow$ I           & 82.5          & 81.6          & 82.5          & 84.7         & 83.0      &  84.1  & 82.6   &    83.9    & 84.6 & 84.9           & \textbf{84.9} \\
E $\rightarrow$ B           & 84.7          & 83.1          & 85.0          & 85.8         & 85.4    &   85.5   & 84.4    &    85.1   & 85.7 & 85.8           & \textbf{86.1} \\
E $\rightarrow$ D           & 84.4          & 81.8          & 85.4          & 86.1         & 85.2     &   86.4  & 83.8      &  84.7   & 86.6 & 86.9           & \textbf{87.3} \\
E $\rightarrow$ K           & 90.2          & 88.4          & 90.8          & 90.9         & 90.7      &  90.8  & 89.7      &   90.3  & 90.1 & 90.8           & \textbf{91.1} \\
E $\rightarrow$ A           & 83.8          & 85.2          & 86.1          & 87.1         & 86.6      &  87.3  & 83.5    &   85.6    & 86.8 & 87.2           & \textbf{87.5} \\
E $\rightarrow$ I           & 79.5          & 79.7          & 80.4          & 82.5         & 81.1      &  82.7  & 79.6   &    81.9    & 82.4 & 82.8           & \textbf{82.8} \\
K $\rightarrow$ B           & 85.1          & 82.6          & 85.3          & 86.1         & 84.9     &   86.5  & 84.8  &    86.1     & 86.3 & 86.5           & \textbf{86.8} \\
K $\rightarrow$ D           & 83.0          & 82.3          & 84.0          & 84.4         & 83.5      &  84.2  & 83.2  &    84.2     & 83.9 & 84.1           & \textbf{84.5} \\
K $\rightarrow$ E           & 88.1          & 86.9          & 88.1          & 88.6         & 88.1     &   88.9  & 87.9  &   88.5      & 89.0 & 89.7           & \textbf{89.7} \\
K $\rightarrow$ A           & 80.3          & 83.0          & 83.9          & 85.7         & 85.8     &   85.6  & 80.2     &   83.2   & 85.8 & 86.0           & \textbf{86.2} \\
K $\rightarrow$ I           & 80.7          & 77.5          & 80.7          & 80.6         & 80.2     &  80.8   & 80.4   &   80.6     & 80.9 & 81.2           & \textbf{81.3} \\
A $\rightarrow$ B           & 76.9          & 77.2          & 78.2          & 78.9         & 78.5     &   78.4  & 77.1  &     78.8    & 79.1 & 79.4           & \textbf{79.6} \\
A $\rightarrow$ D           & 77.6          & 77.9          & 79.4          & 80.2         & 79.8     &   80.0  & 77.4   &    79.2    & 80.0 & 80.1           & \textbf{80.4} \\
A $\rightarrow$ E           & 84.4          & 84.1          & 84.8          & 85.8         & 85.1     &   85.9  & 84.5    &   86.0    & 86.1 & 85.8           & \textbf{86.2} \\
A $\rightarrow$ K           & 85.3          & 82.3          & 85.5          & 87.7         & 85.8     &  87.4   & 84.9   &   86.3     & 87.9 & 88.3           & \textbf{88.3} \\
A $\rightarrow$ I           & 72.1          & 74.9          & 75.4          & 76.4         & 75.2     &   76.1  & 71.8     &   75.7   & 76.2 & 76.7           & \textbf{77.3} \\
I $\rightarrow$ B           & 84.1          & 82.5          & 84.1          & 87.3         & 86.6     &   87.8  & 84.3    &   86.4    & 86.9 & 87.9           & \textbf{88.4} \\
I $\rightarrow$ D           & 84.8          & 83.8          & 84.6          & \textbf{86.5}& 85.7      &  86.2  & 84.6   &    85.8    & 86.2 & 86.3           & \textbf{86.5} \\
I $\rightarrow$ E           & 81.9          & 84.0          & 84.9          & 87.6         & 87.1      &  87.5  & 82.1  &     85.3    & 87.8 & 88.2           & \textbf{88.2} \\
I $\rightarrow$ K           & 85.2          & 84.7          & 85.5          & 85.4         & 85.0     &   84.9  & 85.2     &   85.7   & 85.5 & 85.7           & \textbf{86.0} \\
I $\rightarrow$ A          
           & 82.4          & 83.5               & 84.2          & 85.1        & 84.5    &  85.5    & 82.6           & 84.9 &
 85.4 & 85.4           & \textbf{85.9} \\ \hline
\textbf{Average}            & 83.3          & 82.7               & 84.3          & 85.5         & 84.7     &  85.4   & 83.1     &84.8      & 85.5 & 85.8           & \textbf{86.2} \\ \hline
\textbf{Avg. Spd}         & 1.00$\times$          & 1.00$\times$                & 1.00$\times$          & 1.00$\times$         & 1.00$\times$  &      1.00$\times$   & 1.33$\times$     &   1.53$\times$   & 1.29$\times$ & 1.00$\times$           & \textbf{1.61$\times$} \\ \hline
\end{tabular}
\caption{Main results: This table shows the results of our method compared with other baselines. UDA is the UDApter baseline and CBRT is CeeBERT. Avg. Spd is the average speedup.}
\label{tab: main res}
\end{table*}

\subsection{Experimental Results}
In Tables \ref{tab: main res} and \ref{tab: roberta_res}, we present comprehensive results using the BERT/RoBERTa backbone across various domain pairs for sentiment analysis, while Table \ref{tab: res_nli} showcases results for NLI and entailment classification tasks on BERT. Each experiment is performed five times with different seeds, and the average results are reported. Our method consistently outperforms existing baselines.

While PABEE, an early exiting method without adaptation, shows comparable performance to BERT, both lack domain adaptation techniques. Previous methods like DANN and IATN rely on word2vec or GloVe embeddings, which fail to capture the nuanced word characteristics that BERT’s contextual embeddings can. By leveraging BERT embeddings, we enhance these methods to achieve comparable performance to more advanced baselines.
AAD, which uses BERT but focuses only on the final layer for adaptation, fails to enforce domain-invariant representations across all layers, resulting in suboptimal performance. DAAT improves on prior baselines by incorporating an additional post-training step that makes it domain-aware before performing adversarial training. However, DANN’s method is specific to BERT and not easily extensible.

The performance of DDA suffers from overfitting to the source domain data and does not achieve better speedup or performance compared to our approach due to poor generalization as explained in \cite{ryu2022knowledge}. Additionally, DDA only has a feature classifier to reduce the domain gap which might not be sufficient to effectively minimize the domain discrepancy.

\begin{figure*}[ht]
    \centering
    \begin{subfigure}{0.33\textwidth}
    \centering
    \includegraphics[width=\textwidth]{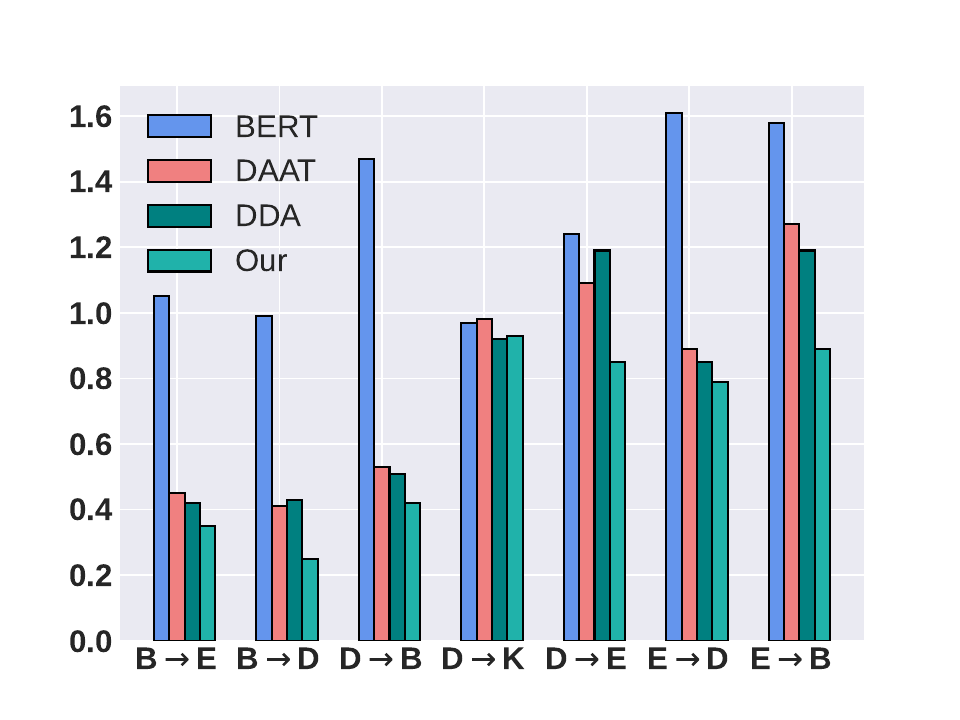}
    \caption{$\mathcal{A}$-distance comparison}
    \label{fig:A-distance}
\end{subfigure}
    \begin{subfigure}{0.299\textwidth}
        \includegraphics[width=\textwidth]{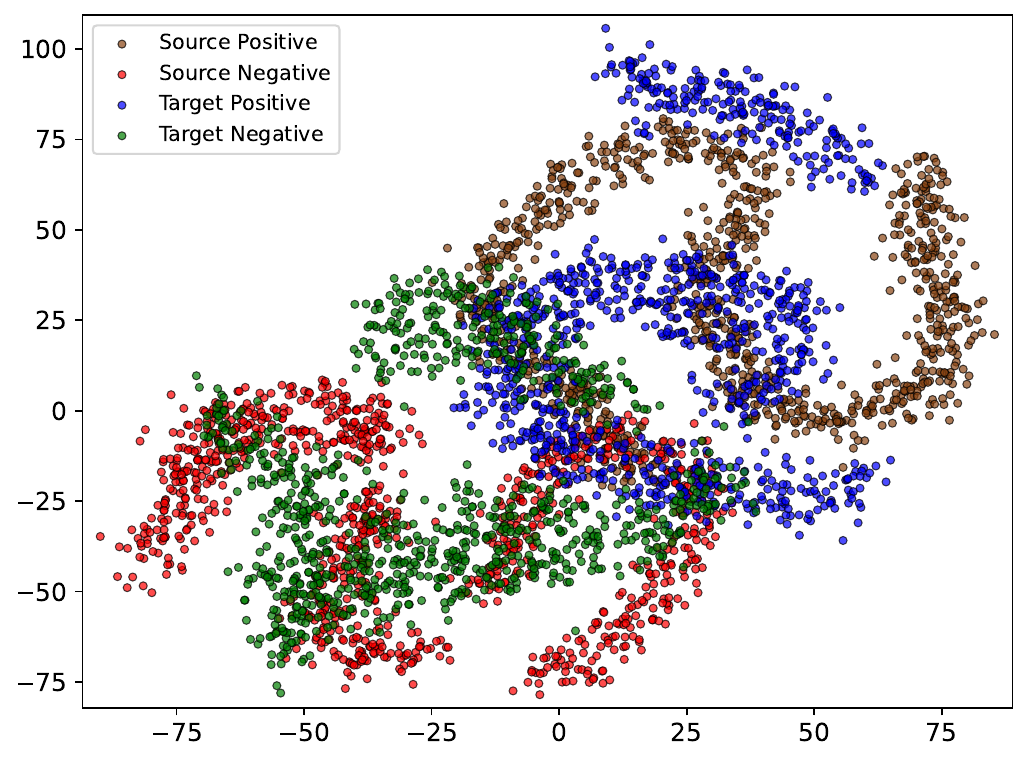}
        \caption{t-SNE visualization of DDA}
        \label{fig:t-sne aad}
    \end{subfigure}
    \begin{subfigure}{0.299\textwidth}
        \includegraphics[width=\textwidth]{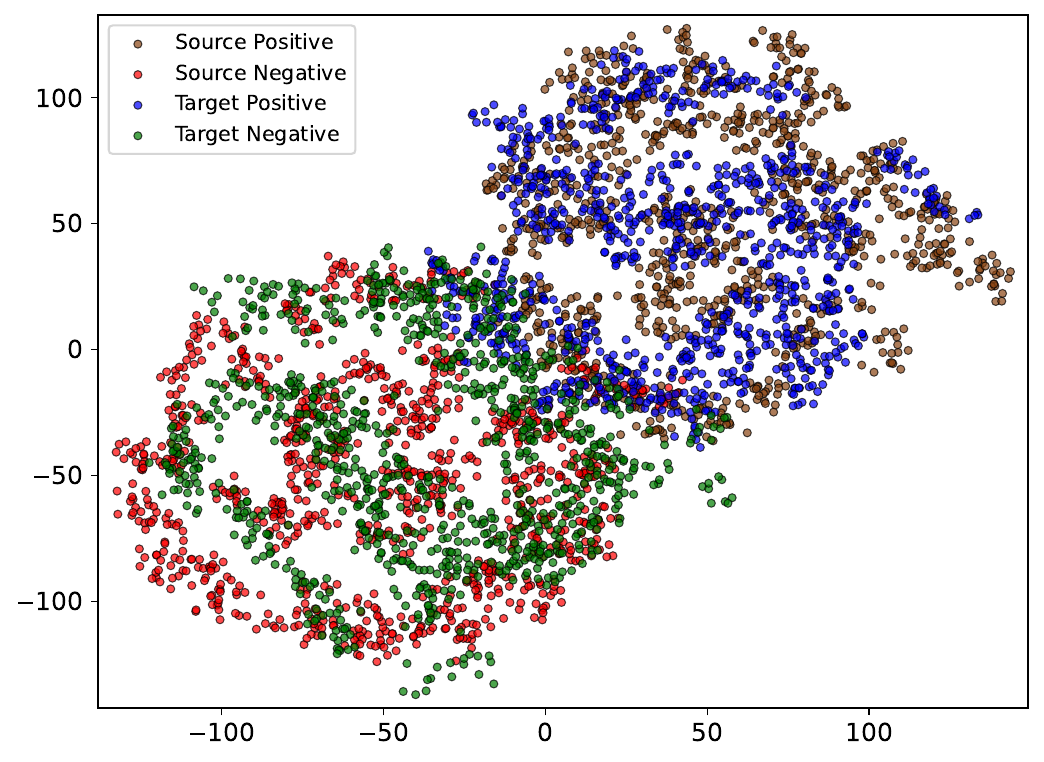}
        \caption{t-SNE visualization of \algo{}}
        \label{fig:t-sne our}
    \end{subfigure}
    \caption{\textbf{Left} figure shows $\mathcal{A}$-distance comparison of our method with different baselines. \textbf{Center and Right} figures show the t-SNE plots for DDA and \algo{} respectively for D $\rightarrow$ E task.}
    \label{fig:global}
\end{figure*}

Our method surpasses previous baselines by adapting to the target domain across all network layers, creating domain-invariant hidden representations throughout rather than relying solely on the final layer. \algo{} effectively reduces the issues of catastrophic forgetting using knowledge distillation which further makes adaptation robust. As shown in the last two columns of Table \ref{tab: main res}, we observe performance improvements even with early predictions, mitigating the issue of overthinking.

Moreover, we achieve an average speedup of 1.61x, with the highest at 2.11x and the lowest at 1.00x, demonstrating our method's efficiency in both domain adaptation and faster inference, making the model both quick and robust.

\subsection{$\mathcal{A}$-distance}

From Equation \ref{eq: distance}, we note the significance of domain divergence, a pivotal metric in assessing the efficacy of domain adaptation methods. To quantify this, we calculate the $\mathcal{A}$-distance, commonly utilized for measuring domain dissimilarity. Defined as $d_{\mathcal{A}} = 2(1-2\epsilon)$, where $\epsilon$ denotes the generalization error of a classifier trained to discern source from target domain samples. Following existing methods, we divide the data into equal subsets (both source and target), train a linear SVM on one subset to classify domain origin, and evaluate error rates on the other subset, deriving the $\mathcal{A}$-distance accordingly.

Comparing the $\mathcal{A}$-distance across BERT, DAAT, DDA, and our method as these are the top best-performing methods, BERT consistently exhibits the highest $\mathcal{A}$-distance across dataset pairs. DDA and DAAT demonstrate lower $\mathcal{A}$-distance, attributed to their application of adversarial training, which mitigates domain dissimilarity. Our method achieves the lowest $\mathcal{A}$-distance, owing to adversarial training and reduction of catastrophic forgetting across all layers.

\begin{table}[ht]
\centering
\small
\begin{tabular}{cccccc}
\hline
\textbf{S $\rightarrow$ T}      & \textbf{RBERT} & \textbf{AAD} & \textbf{DAAT} & \textbf{final} & \textbf{Our}  \\ \hline
B $\rightarrow$ D               & 87.9          & 88.1         & 88.5          & 88.7               & \textbf{89.0} \\
B $\rightarrow$ E               & 89.5          & 89.6         & 89.2          & 90.4               & \textbf{90.4} \\
B $\rightarrow$ K               & 90.4          & 92.1         & 91.9          & 92.5               & \textbf{92.8} \\
B $\rightarrow$ A               & 84.6          & 85.9         & 86.3          & 86.9               & \textbf{86.9} \\
B $\rightarrow$ I               & 85.7          & 85.5         & 85.9          & 86.2               & \textbf{86.3} \\
D $\rightarrow$ B               & 87.2          & 89.4         & 89.7          & 90.1               & \textbf{90.3} \\
D $\rightarrow$ E               & 89.6          & 89.8         & 90.1          & 90.4               & \textbf{90.5} \\
D $\rightarrow$ K               & 88.5          & 91.1         & 92.0            & 91.9               & \textbf{92.1} \\
D $\rightarrow$ A               & 84.8          & 85.6         & 85.5          & 86.2               & \textbf{86.2} \\
D $\rightarrow$ I               & 86.3          & 86.4         & 86.9          & 87.3               & \textbf{87.8} \\
E $\rightarrow$ B               & 85.9          & 88.2         & 88.7          & 89.8               & \textbf{90.0} \\
E $\rightarrow$ D               & 84.1          & 87.8         & 87.3          & 88.5               & \textbf{88.8} \\
E $\rightarrow$ K               & 92.4          & 92.9         & 92.6          & 93.4               & \textbf{93.4} \\
E $\rightarrow$ A               & 84.0          & 87.1         & 86.4          & 87.7               & \textbf{87.9} \\
E $\rightarrow$ I               & 79.4          & 84.9         & 83.7          & 85.4               & \textbf{85.6} \\ \hline
\textbf{Average}     & 86.7          & 88.3         & 88.3          & 89.0               & \textbf{89.2} \\ \hline
\textbf{Avg.Speed} & 1.00          & 1.00         & 1.00          & 1.00               & \textbf{1.67} \\ \hline
\end{tabular}
\caption{This method shows the results of \algo{} and other baselines on the RoBERTa-base(RBERT) model.}
\label{tab: roberta_res}
\end{table}
\subsection{Feature visualization}
For an intuitive understanding of the effects of multi-level domain adaptation on BERT, we further perform visualization of feature representations of our method and the DDA method. We perform the feature representation visualization on the test split of the source and target domain for the D $\rightarrow$ E task. In figure \ref{fig:t-sne aad}, \ref{fig:t-sne our}, we apply t-SNE (t-distributed Stochastic Neighbor Embedding) on the set of all representations of source and target data points. Every sample is mapped into a 768-dimensional feature space by BERT and projected back to the two-dimensional plane by the t-SNE.
\vspace{-0.09cm}

From figure \ref{fig:t-sne aad}, we can observe that there is a lack of distinction between source positive and source negative samples in DDA and many points overlap when in target positive and target negatives. Still, it manages to distinguish between the larger portion of data from the target domain due to adaptive inference. For our method, the target domain positives and negatives are well-separated with few overlaps showing that the multi-level domain adaptation in an adversarial setup can benefit in bridging the domain gap more efficiently. Also, in our method, there is a better overlap between source positives, target positives, and source negatives, as well as target negatives, which further shows that our method can give better domain invariant features.   

\begin{table}[ht]
\small
\begin{tabular}{cccccc}
\hline
\textbf{S to T}    & \textbf{BERT} & \textbf{CBRT} & \textbf{DDA} & \textbf{Final} & \textbf{Our}  \\ \hline
Mn to Sn           & 80.5          & 79.8          & 81.8         & 83.1           & \textbf{83.3} \\
Sn to Mn           & 86.9          & 87.1          & 88.5         & 89.3           & \textbf{89.3} \\
R to Q             & 67.2          & 68.5          & 69.2         & 71.0           & \textbf{71.1} \\
Q to R             & 65.8          & 66.3          & 67.9         & 68.4           & \textbf{68.7} \\
Mr to Sc           & 79.5          & 80.1          & 81.5         & 82.9           & \textbf{83.0}   \\
Sc to Mr           & 85.3          & 85.7          & 86.3         & 87.6           & \textbf{87.6} \\ \hline
\textbf{Average}   & 77.5          & 77.9          & 79.2         & 80.3           & \textbf{80.5} \\ \hline
\textbf{Avg Speed} & 1.00          & 1.68          & 1.32         & 1.00           & \textbf{1.71} \\ \hline
\end{tabular}
\caption{Results on NLI and entailment classification tasks. The abbreviations are MNLI (Mn), SNLI (Sn), RTE (R), QQP (Q), MRPC (M) and SciTail (Sc).}
\label{tab: res_nli}
\end{table}

\begin{table}[ht]
\centering
\small
\begin{tabular}{cccccc}
\hline
\textbf{S $\rightarrow$ T} & \textbf{\#} & \textbf{BERT} & \textbf{DAAT} & \textbf{UDA} & \textbf{Our}  \\ \hline
B $\rightarrow$ D          & 500          & 79.0          & 81.5         & 82.5          & \textbf{85.8$\pm$1.0} \\
D $\rightarrow$ B          & 500          & 75.6          & 81.8         & 82.1          & \textbf{85.1$\pm$0.7} \\
B $\rightarrow$ D          & 1000          & 81.5          & 82.0         & 83.3          & \textbf{86.9$\pm$0.6} \\
D $\rightarrow$ B          & 1000          & 80.1          & 83.4         & 83.7          & \textbf{86.5$\pm$0.5} \\
B $\rightarrow$ D          & 1500          & 84.5          & 86.9         & 87.1          & \textbf{88.2$\pm$0.5} \\
D $\rightarrow$ B          & 1500          & 84.2          & 86.7         & 87.0          & \textbf{87.9$\pm$0.2} \\ \hline
\end{tabular}
\caption{This table shows the scalability as well as stability results of our method. \# shows the number of samples of source dataset used to train the backbone.}
\label{tab: scalability}
\vspace{-0.39cm}
\end{table}

\subsection{Ablation study}
In this section, we show the robustness of our method to source dataset size and its stability. The effect of changing the hyperparameter $\alpha$ is given in the Appendix \ref{sec: speedup}.

In Table \ref{tab: scalability}, we demonstrate the scalability of our method by varying the size of the source dataset used for training. We use different fractions of samples as training data for the source backbone. Our findings highlight a key strength of our approach: robustness to variations in the size of the source dataset. As shown in Table \ref{tab: scalability}, reducing the size of the labeled set significantly impacts other models' performance, whereas our method experiences only a slight performance decline. This robustness is attributed to the multi-level domain adaptation integrated into our framework through early exits, effectively utilizing limited labeled data and incorporating valuable representations from the training dataset. The distillation loss applied at every layer also enhances robustness, as previous methods often only attach knowledge distillation to the final layer, potentially leading to catastrophic forgetting or mode collapse. Our approach addresses this by mitigating noise accumulation across all layers.

Moreover, early exit models reduce the chances of overthinking which further helps in better results when the labeled dataset size varies as reflected in our results of table \ref{tab: scalability}.

We also assess the stability of our method by calculating the standard deviation across five random runs with different seeds. Early exiting models have shown good generalization capabilities in previous works \cite{zhu2021leebert, zhou2020bert}, further verifying the scalability and stability of our method. The stability performance of other methods is provided in Table \ref{tab: scalability_all} in the Appendix.
\section{Conclusion}
We present a new method \algo{} for multi-level domain adaptation in PLMs using the early exit approach and adversarial training. The exits attached lower the chances of catastrophic forgetting by incorporating knowledge distillation across all layers. Also, the attached exits decrease the inference time by inferring the easier samples early. t-SNE plots and $\mathcal{A}$-distance demonstrate the effectiveness of our method in domain adaptation. Extensive experiments demonstrate that our method not only bridges the domain gap in a cross-domain setup but also provides faster inference making it suitable for real-world applications.

\section{Limitations}
Our model uses knowledge distillation at every layer which is a sensitive part of the method if we remove the knowledge distillation loss from a few layers, it might lead to noise accumulation from those layers as each layer is forced to give similar representations as the target domain. If a few layers face the mode collapse, it might lead to degraded performance.

Also, after adapting to the target domain, we can adapt to the threshold values based on the confidence values given by the exits at each layer so that the inference at each layer becomes more effective.

\section*{Acknowledgements}
Divya Jyoti Bajpai is supported by the Prime Minister’s Research Fellowship (PMRF), Govt. of India.  Manjesh K. Hanawal thanks funding support from SERB, Govt. of India, through the Core Research Grant (CRG/2022/008807) and MATRICS grant (MTR/2021/000645), and DST-Inria Targeted Programme.

\bibliography{anthology,custom}

\begin{thebibliography}{44}
\expandafter\ifx\csname natexlab\endcsname\relax\def\natexlab#1{#1}\fi

\bibitem[{Ajakan et~al.(2014)Ajakan, Germain, Larochelle, Laviolette, and
  Marchand}]{ajakan2014domain}
Hana Ajakan, Pascal Germain, Hugo Larochelle, Fran{\c{c}}ois Laviolette, and
  Mario Marchand. 2014.
\newblock Domain-adversarial neural networks.
\newblock \emph{arXiv preprint arXiv:1412.4446}.

\bibitem[{Bajpai and Hanawal(2024)}]{bajpai2024ceebert}
Divya~Jyoti Bajpai and Manjesh~Kumar Hanawal. 2024.
\newblock Ceebert: Cross-domain inference in early exit bert.
\newblock \emph{arXiv preprint arXiv:2405.15039}.

\bibitem[{Banino et~al.(2021)Banino, Balaguer, and
  Blundell}]{banino2021pondernet}
Andrea Banino, Jan Balaguer, and Charles Blundell. 2021.
\newblock Pondernet: Learning to ponder.
\newblock \emph{arXiv preprint arXiv:2107.05407}.

\bibitem[{Bapna et~al.(2020)Bapna, Arivazhagan, and
  Firat}]{bapna2020controlling}
Ankur Bapna, Naveen Arivazhagan, and Orhan Firat. 2020.
\newblock Controlling computation versus quality for neural sequence models.
\newblock \emph{arXiv preprint arXiv:2002.07106}.

\bibitem[{Ben-David et~al.(2010)Ben-David, Blitzer, Crammer, Kulesza, Pereira,
  and Vaughan}]{ben2010theory}
Shai Ben-David, John Blitzer, Koby Crammer, Alex Kulesza, Fernando Pereira, and
  Jennifer~Wortman Vaughan. 2010.
\newblock A theory of learning from different domains.
\newblock \emph{Machine learning}, 79:151--175.

\bibitem[{Blitzer et~al.(2007)Blitzer, Dredze, and
  Pereira}]{blitzer2007biographies}
John Blitzer, Mark Dredze, and Fernando Pereira. 2007.
\newblock Biographies, bollywood, boom-boxes and blenders: Domain adaptation
  for sentiment classification.
\newblock In \emph{Proceedings of the 45th annual meeting of the association of
  computational linguistics}, pages 440--447.

\bibitem[{Bousmalis et~al.(2016)Bousmalis, Trigeorgis, Silberman, Krishnan, and
  Erhan}]{bousmalis2016domain}
Konstantinos Bousmalis, George Trigeorgis, Nathan Silberman, Dilip Krishnan,
  and Dumitru Erhan. 2016.
\newblock Domain separation networks.
\newblock \emph{Advances in neural information processing systems}, 29.

\bibitem[{Devlin et~al.(2018)Devlin, Chang, Lee, and
  Toutanova}]{devlin2018bert}
Jacob Devlin, Ming-Wei Chang, Kenton Lee, and Kristina Toutanova. 2018.
\newblock Bert: Pre-training of deep bidirectional transformers for language
  understanding.
\newblock \emph{arXiv preprint arXiv:1810.04805}.

\bibitem[{Du et~al.(2020)Du, Sun, Wang, Qi, and Liao}]{du2020adversarial}
Chunning Du, Haifeng Sun, Jingyu Wang, Qi~Qi, and Jianxin Liao. 2020.
\newblock Adversarial and domain-aware bert for cross-domain sentiment
  analysis.
\newblock In \emph{Proceedings of the 58th annual meeting of the Association
  for Computational Linguistics}, pages 4019--4028.

\bibitem[{Elbayad et~al.(2020)Elbayad, Gu, Grave, and Auli}]{elbayad19arxiv}
Maha Elbayad, Jiatao Gu, Edouard Grave, and Michael Auli. 2020.
\newblock Depth-adaptive transformer.
\newblock In \emph{In Proc. of ICLR}.

\bibitem[{Fei et~al.(2022)Fei, Yan, Wang, and Tian}]{fei2022deecap}
Zhengcong Fei, Xu~Yan, Shuhui Wang, and Qi~Tian. 2022.
\newblock Deecap: Dynamic early exiting for efficient image captioning.
\newblock In \emph{Proceedings of the IEEE/CVF Conference on Computer Vision
  and Pattern Recognition}, pages 12216--12226.

\bibitem[{He et~al.(2021)He, Keivanloo, Xu, He, Zeng, Rajagopalan, and
  Chilimbi}]{he2021magic}
Xuanli He, Iman Keivanloo, Yi~Xu, Xiang He, Belinda Zeng, Santosh Rajagopalan,
  and Trishul Chilimbi. 2021.
\newblock Magic pyramid: Accelerating inference with early exiting and token
  pruning.
\newblock \emph{arXiv preprint arXiv:2111.00230}.

\bibitem[{Hoffman et~al.(2018)Hoffman, Tzeng, Park, Zhu, Isola, Saenko, Efros,
  and Darrell}]{hoffman2018cycada}
Judy Hoffman, Eric Tzeng, Taesung Park, Jun-Yan Zhu, Phillip Isola, Kate
  Saenko, Alexei Efros, and Trevor Darrell. 2018.
\newblock Cycada: Cycle-consistent adversarial domain adaptation.
\newblock In \emph{International conference on machine learning}, pages
  1989--1998. Pmlr.

\bibitem[{Huang et~al.(2017)Huang, Chen, Li, Wu, Van Der~Maaten, and
  Weinberger}]{huang2017multi}
Gao Huang, Danlu Chen, Tianhong Li, Felix Wu, Laurens Van Der~Maaten, and
  Kilian~Q Weinberger. 2017.
\newblock Multi-scale dense networks for resource efficient image
  classification.
\newblock \emph{arXiv preprint arXiv:1703.09844}.

\bibitem[{Ji et~al.(2023)Ji, Wang, Li, Chen, Chen, and Zhang}]{ji2023early}
Yixin Ji, Jikai Wang, Juntao Li, Qiang Chen, Wenliang Chen, and Min Zhang.
  2023.
\newblock Early exit with disentangled representation and equiangular tight
  frame.
\newblock In \emph{Findings of the Association for Computational Linguistics:
  ACL 2023}, pages 14128--14142.

\bibitem[{Jiao et~al.(2019)Jiao, Yin, Shang, Jiang, Chen, Li, Wang, and
  Liu}]{jiao2019tinybert}
Xiaoqi Jiao, Yichun Yin, Lifeng Shang, Xin Jiang, Xiao Chen, Linlin Li, Fang
  Wang, and Qun Liu. 2019.
\newblock Tinybert: Distilling bert for natural language understanding.
\newblock \emph{arXiv preprint arXiv:1909.10351}.

\bibitem[{Kaya et~al.(2019)Kaya, Hong, and Dumitras}]{kaya2019shallow}
Yigitcan Kaya, Sanghyun Hong, and Tudor Dumitras. 2019.
\newblock Shallow-deep networks: Understanding and mitigating network
  overthinking.
\newblock In \emph{International conference on machine learning}, pages
  3301--3310. PMLR.

\bibitem[{Kim et~al.(2021)Kim, Gholami, Yao, Mahoney, and
  Keutzer}]{kim2021bert}
Sehoon Kim, Amir Gholami, Zhewei Yao, Michael~W Mahoney, and Kurt Keutzer.
  2021.
\newblock I-bert: Integer-only bert quantization.
\newblock In \emph{International conference on machine learning}, pages
  5506--5518. PMLR.

\bibitem[{Kingma and Ba(2014)}]{kingma2014adam}
Diederik~P Kingma and Jimmy Ba. 2014.
\newblock Adam: A method for stochastic optimization.
\newblock \emph{arXiv preprint arXiv:1412.6980}.

\bibitem[{Li et~al.(2021)Li, Zhang, Ma, Liu, and Li}]{li2021dynamic}
Shuang Li, Jinming Zhang, Wenxuan Ma, Chi~Harold Liu, and Wei Li. 2021.
\newblock Dynamic domain adaptation for efficient inference.
\newblock In \emph{Proceedings of the IEEE/CVF conference on computer vision
  and pattern recognition}, pages 7832--7841.

\bibitem[{Liu et~al.(2021)Liu, Sun, He, Wu, Zhang, Jiang, Cao, Huang, and
  Qiu}]{liu2021elasticbert}
Xiangyang Liu, Tianxiang Sun, Junliang He, Lingling Wu, Xinyu Zhang, Hao Jiang,
  Zhao Cao, Xuanjing Huang, and Xipeng Qiu. 2021.
\newblock \href {https://arxiv.org/abs/2110.07038} {Towards efficient {NLP:}
  {A} standard evaluation and {A} strong baseline}.

\bibitem[{Liu et~al.(2019)Liu, Ott, Goyal, Du, Joshi, Chen, Levy, Lewis,
  Zettlemoyer, and Stoyanov}]{liu2019roberta}
Yinhan Liu, Myle Ott, Naman Goyal, Jingfei Du, Mandar Joshi, Danqi Chen, Omer
  Levy, Mike Lewis, Luke Zettlemoyer, and Veselin Stoyanov. 2019.
\newblock Roberta: A robustly optimized bert pretraining approach.
\newblock \emph{arXiv preprint arXiv:1907.11692}.

\bibitem[{Long et~al.(2015)Long, Cao, Wang, and Jordan}]{long2015learning}
Mingsheng Long, Yue Cao, Jianmin Wang, and Michael Jordan. 2015.
\newblock Learning transferable features with deep adaptation networks.
\newblock In \emph{International conference on machine learning}, pages
  97--105. PMLR.

\bibitem[{Maas et~al.(2011)Maas, Daly, Pham, Huang, Ng, and
  Potts}]{maas2011learning}
Andrew Maas, Raymond~E Daly, Peter~T Pham, Dan Huang, Andrew~Y Ng, and
  Christopher Potts. 2011.
\newblock Learning word vectors for sentiment analysis.
\newblock In \emph{Proceedings of the 49th annual meeting of the association
  for computational linguistics: Human language technologies}, pages 142--150.

\bibitem[{Malik et~al.(2023)Malik, Kashyap, Kan, and Poria}]{malik2023udapter}
Bhavitvya Malik, Abhinav~Ramesh Kashyap, Min-Yen Kan, and Soujanya Poria. 2023.
\newblock Udapter--efficient domain adaptation using adapters.
\newblock \emph{arXiv preprint arXiv:2302.03194}.

\bibitem[{Michel et~al.(2019)Michel, Levy, and Neubig}]{michel2019sixteen}
Paul Michel, Omer Levy, and Graham Neubig. 2019.
\newblock Are sixteen heads really better than one?
\newblock \emph{Advances in neural information processing systems}, 32.

\bibitem[{Nguyen(2015)}]{nguyen2015airline}
Quang Nguyen. 2015.
\newblock The airline review dataset.
\newblock \emph{Scraped from www. airlinequality. com, https://github.
  com/quankiquanki/skytrax-reviews-dataset}.

\bibitem[{Peng et~al.(2018)Peng, Zhang, Jiang, and Huang}]{peng2018cross}
Minlong Peng, Qi~Zhang, Yu-gang Jiang, and Xuan-Jing Huang. 2018.
\newblock Cross-domain sentiment classification with target domain specific
  information.
\newblock In \emph{Proceedings of the 56th Annual Meeting of the Association
  for Computational Linguistics (Volume 1: Long Papers)}, pages 2505--2513.

\bibitem[{Radford et~al.(2019)Radford, Wu, Child, Luan, Amodei, Sutskever
  et~al.}]{radford2019language}
Alec Radford, Jeffrey Wu, Rewon Child, David Luan, Dario Amodei, Ilya
  Sutskever, et~al. 2019.
\newblock Language models are unsupervised multitask learners.
\newblock \emph{OpenAI blog}, 1(8):9.

\bibitem[{Ryu et~al.(2022)Ryu, Lee, and Lee}]{ryu2022knowledge}
Minho Ryu, Geonseok Lee, and Kichun Lee. 2022.
\newblock Knowledge distillation for bert unsupervised domain adaptation.
\newblock \emph{Knowledge and Information Systems}, 64(11):3113--3128.

\bibitem[{Sun and Saenko(2016)}]{sun2016deep}
Baochen Sun and Kate Saenko. 2016.
\newblock Deep coral: Correlation alignment for deep domain adaptation.
\newblock In \emph{Computer Vision--ECCV 2016 Workshops: Amsterdam, The
  Netherlands, October 8-10 and 15-16, 2016, Proceedings, Part III 14}, pages
  443--450. Springer.

\bibitem[{Tang et~al.(2023)Tang, Wang, Kong, Zhang, Li, Ding, Wang, Liang, and
  Xu}]{tang2023you}
Shengkun Tang, Yaqing Wang, Zhenglun Kong, Tianchi Zhang, Yao Li, Caiwen Ding,
  Yanzhi Wang, Yi~Liang, and Dongkuan Xu. 2023.
\newblock You need multiple exiting: Dynamic early exiting for accelerating
  unified vision language model.
\newblock In \emph{Proceedings of the IEEE/CVF Conference on Computer Vision
  and Pattern Recognition}, pages 10781--10791.

\bibitem[{Teerapittayanon et~al.(2016)Teerapittayanon, McDanel, and
  Kung}]{teerapittayanon2016branchynet}
Surat Teerapittayanon, Bradley McDanel, and Hsiang-Tsung Kung. 2016.
\newblock Branchynet: Fast inference via early exiting from deep neural
  networks.
\newblock In \emph{2016 23rd International Conference on Pattern Recognition
  (ICPR)}, pages 2464--2469. IEEE.

\bibitem[{Tzeng et~al.(2017)Tzeng, Hoffman, Saenko, and
  Darrell}]{tzeng2017adversarial}
Eric Tzeng, Judy Hoffman, Kate Saenko, and Trevor Darrell. 2017.
\newblock Adversarial discriminative domain adaptation.
\newblock In \emph{Proceedings of the IEEE conference on computer vision and
  pattern recognition}, pages 7167--7176.

\bibitem[{Tzeng et~al.(2014)Tzeng, Hoffman, Zhang, Saenko, and
  Darrell}]{tzeng2014deep}
Eric Tzeng, Judy Hoffman, Ning Zhang, Kate Saenko, and Trevor Darrell. 2014.
\newblock Deep domain confusion: Maximizing for domain invariance.
\newblock \emph{arXiv preprint arXiv:1412.3474}.

\bibitem[{Wang et~al.(2019)Wang, Singh, Michael, Hill, Levy, and
  Bowman}]{wang2019glue}
Alex Wang, Amanpreet Singh, Julian Michael, Felix Hill, Omer Levy, and
  Samuel~R. Bowman. 2019.
\newblock {GLUE}: A multi-task benchmark and analysis platform for natural
  language understanding.
\newblock In the Proceedings of ICLR.

\bibitem[{Xin et~al.(2020)Xin, Tang, Lee, Yu, and Lin}]{xin2020deebert}
Ji~Xin, Raphael Tang, Jaejun Lee, Yaoliang Yu, and Jimmy Lin. 2020.
\newblock Deebert: Dynamic early exiting for accelerating bert inference.
\newblock \emph{arXiv preprint arXiv:2004.12993}.

\bibitem[{Xin et~al.(2021)Xin, Tang, Yu, and Lin}]{xin2021berxit}
Ji~Xin, Raphael Tang, Yaoliang Yu, and Jimmy Lin. 2021.
\newblock Berxit: Early exiting for bert with better fine-tuning and extension
  to regression.
\newblock In \emph{Proceedings of the 16th conference of the European chapter
  of the association for computational linguistics: Main Volume}, pages
  91--104.

\bibitem[{Yang et~al.(2019)Yang, Dai, Yang, Carbonell, Salakhutdinov, and
  Le}]{yang2019xlnet}
Zhilin Yang, Zihang Dai, Yiming Yang, Jaime Carbonell, Russ~R Salakhutdinov,
  and Quoc~V Le. 2019.
\newblock Xlnet: Generalized autoregressive pretraining for language
  understanding.
\newblock \emph{Advances in neural information processing systems}, 32.

\bibitem[{Yu and Jiang(2016)}]{yu2016learning}
Jianfei Yu and Jing Jiang. 2016.
\newblock Learning sentence embeddings with auxiliary tasks for cross-domain
  sentiment classification.
\newblock In \emph{Proceedings of the 2016 conference on empirical methods in
  natural language processing}, pages 236--246.

\bibitem[{Zhang et~al.(2019)Zhang, Zhang, Liu, Zhao, Zhu, and
  Chen}]{zhang2019interactive}
Kai Zhang, Hefu Zhang, Qi~Liu, Hongke Zhao, Hengshu Zhu, and Enhong Chen. 2019.
\newblock Interactive attention transfer network for cross-domain sentiment
  classification.
\newblock In \emph{Proceedings of the AAAI Conference on Artificial
  Intelligence}, volume~33, pages 5773--5780.

\bibitem[{Zhou et~al.(2020)Zhou, Xu, Ge, McAuley, Xu, and Wei}]{zhou2020bert}
Wangchunshu Zhou, Canwen Xu, Tao Ge, Julian McAuley, Ke~Xu, and Furu Wei. 2020.
\newblock Bert loses patience: Fast and robust inference with early exit.
\newblock \emph{Advances in Neural Information Processing Systems},
  33:18330--18341.

\bibitem[{Zhu(2021)}]{zhu2021leebert}
Wei Zhu. 2021.
\newblock Leebert: Learned early exit for bert with cross-level optimization.
\newblock In \emph{Proceedings of the 59th Annual Meeting of the Association
  for Computational Linguistics and the 11th International Joint Conference on
  Natural Language Processing (Volume 1: Long Papers)}, pages 2968--2980.

\bibitem[{Ziser and Reichart(2018)}]{ziser2018pivot}
Yftah Ziser and Roi Reichart. 2018.
\newblock Pivot based language modeling for improved neural domain adaptation.
\newblock In \emph{Proceedings of the 2018 Conference of the North American
  Chapter of the Association for Computational Linguistics: Human Language
  Technologies, Volume 1 (Long Papers)}, pages 1241--1251.

\end{thebibliography}
\newpage
\appendix
\section{Appendix}
\label{sec:appendix}
\subsection{Stability}
In table \ref{tab: scalability_all}, we have shown the stability as well as scalability results of all the state-of-the-art baselines, we can observe that the stability of our method is close but slightly better than the previous methods. The reason for better stability is the better generalization achieved using the early exits. 

\subsection{Accuracy vs Speedup}\label{sec: speedup}
\begin{figure}
        \includegraphics[scale = 0.39]{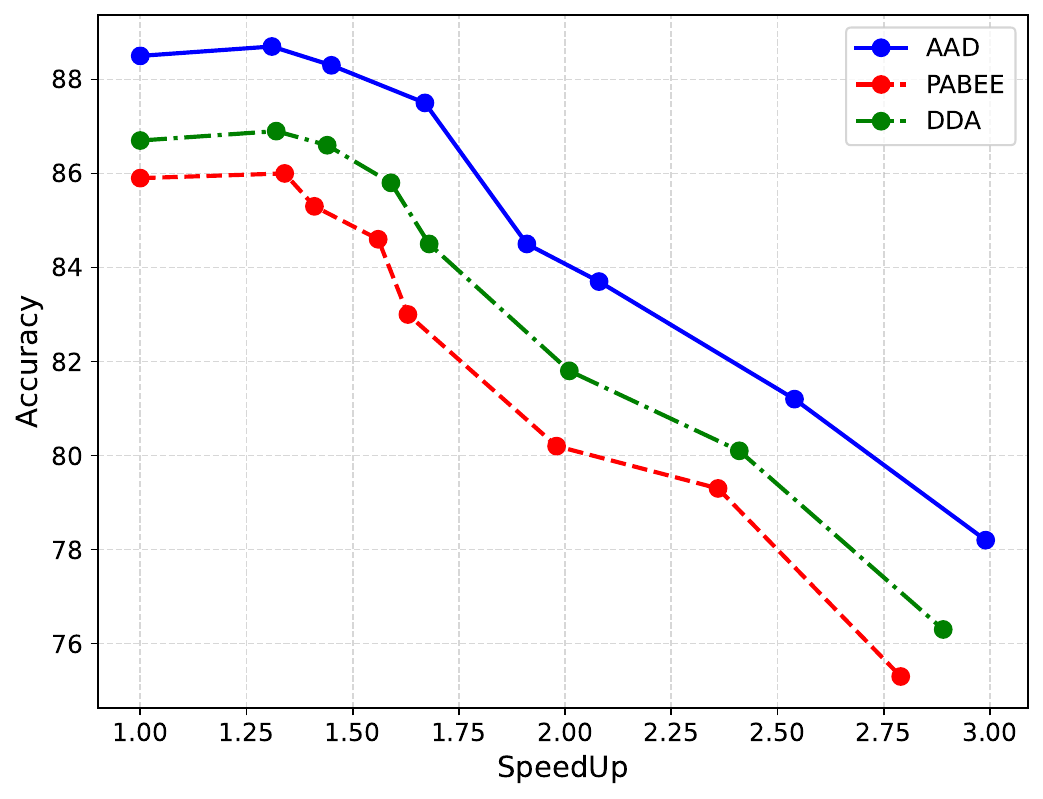}
    \caption{Speedup vs Accuracy}
    \label{fig:speedvsacc}
\end{figure}

In figure \ref{fig:speedvsacc}, we provide the results of changing the values of $\alpha$ used to model the accuracy efficiency trade-off. The higher values of $\alpha$ provide accurate predictions but with lower speedup. We have plotted the PABEE as well as our method which we get by changing the patience parameter $t$. Since PABEE is not adapted the performance is lower than our method. Observe that there is a slight increase in the plot which is due to the overthinking issue faced by the final layer.
\begin{table*}
\centering
\small
\begin{tabular}{cccccc}
\hline
\textbf{S $\rightarrow$ T} & \textbf{\#} & \textbf{BERT} & \textbf{AAD} & \textbf{DAAT} & \textbf{Our}  \\ \hline
B $\rightarrow$ D          & 500          & 81.0$\pm$1.7          & 80.5$\pm$1.1        & 82.7$\pm$1.2          & \textbf{85.1$\pm$1.0} \\
D $\rightarrow$ B          & 500          & 73.6$\pm$1.4          & 81.3$\pm$0.9         & 82.1$\pm$0.9          & \textbf{83.4$\pm$0.7} \\
B $\rightarrow$ D          & 1000          & 79.5$\pm$1.5          & 82.1$\pm$0.6         & 82.5$\pm$0.9          & \textbf{85.3$\pm$0.6} \\
D $\rightarrow$ B          & 1000          & 80.1$\pm$0.9          & 81.4$\pm$0.6         & 83.1$\pm$0.6          & \textbf{85.7$\pm$0.5} \\
B $\rightarrow$ D          & 1500          & 82.5$\pm$0.5          & 83.4$\pm$0.4         & 84.2$\pm$0.7          & \textbf{86.5$\pm$0.5} \\
D $\rightarrow$ B          & 1500          & 82.8$\pm$0.5          & 83.8$\pm$0.3         & 84.2$\pm$0.4          & \textbf{84.7$\pm$0.2} \\ \hline
\end{tabular}
\caption{This table shows the scalability as well as stability results of our method. \# shows the number of samples of source dataset used to train the backbone.}
\label{tab: scalability_all}
\end{table*}

\begin{table}
\centering
\begin{tabular}{|l|l|}
\hline
\textbf{Dataset} & \textbf{\#Samples} \\ \hline
MNLI             & 433K                         \\ \hline
MRPC             & 4K        \\ \hline      
SNLI             & 550K                        \\ \hline
QQP              & 365K                              \\ \hline
SciTail          & 24K                          \\ \hline
RTE              & 2.5K  \\ \hline
\end{tabular}
\caption{This table provides the sizes of the datasets.}
\label{tab: dataset}
\end{table}

\subsection{Dataset statistics and \# Parameters}
The Amazon review dataset consists of $2000$ labeled samples that are used for training and development and then the model is test split consisting of $3000-5000$ samples. For the GLUE and ELUE tasks, the dataset statistics are given in Table \ref{tab: dataset}.

The number of parameters in BERT/RoBERTa-base is 110 Million and for BERT/RoBERTa-Large is 340 Million.

\end{document}